\documentclass[12pt,a4paper]{article}

\usepackage{arxiv}

\usepackage[utf8]{inputenc} % allow utf-8 input
\usepackage[T1]{fontenc}    % use 8-bit T1 fonts
\usepackage{hyperref}       % hyperlinks
\usepackage{url}            % simple URL typesetting
\usepackage{booktabs}       % professional-quality tables
\usepackage{amsfonts}       % blackboard math symbols
\usepackage{nicefrac}       % compact symbols for 1/2, etc.
\usepackage{microtype}      % microtypography
\usepackage{lipsum}
\usepackage{graphicx}
\usepackage{bbm}
\usepackage[caption=false]{subfig}
\usepackage{multirow}
\usepackage{multibib}
\usepackage{siunitx}
\usepackage{tablefootnote}
\newcites{supp}{References}
\graphicspath{ {./images/} }

\title{Optimizing Carbon Storage Operations for Long-Term Safety}

\author{
 Yizheng Wang \\
  Earth and Planetary Sciences\\
  Stanford University\\
  Stanford, CA 94305 \\
  \texttt{yizhengw@stanford.edu} \\
   \And
 Markus Zechner \\
  Earth and Planetary Sciences\\
  Stanford University\\
  Stanford, CA 94305 \\
  \texttt{mzechner@stanford.edu} \\
  \And
 Gege Wen \\
  Energy Science and Engineering\\
  Stanford University\\
  Stanford, CA 94305 \\
  \texttt{gegewen@stanford.edu} \\
\And
Anthony Louis Corso \\
  Aeronautics and Astronautics\\
  Stanford University\\
  Stanford, CA 94305 \\
  \texttt{acorso@stanford.edu} \\
  \And
John Michael Mern \\
  Aeronautics and Astronautics\\
  Stanford University\\
  Stanford, CA 94305 \\
  \texttt{jmern91@stanford.edu} \\
    \And
Mykel J. Kochenderfer \\
  Aeronautics and Astronautics\\
  Stanford University\\
  Stanford, CA 94305 \\
  \texttt{mykel@stanford.edu} \\
      \And
Jef Karel Caers \\
  Earth and Planetary Sciences\\
  Stanford University\\
  Stanford, CA 94305 \\
  \texttt{jcaers@stanford.edu} \\
}

\begin{document}
\maketitle
\begin{abstract}
To combat global warming and mitigate the risks associated with climate change, carbon capture and storage (CCS) has emerged as a crucial technology. However, safely sequestering CO\textsubscript{2} in geological formations for long-term storage presents several challenges. In this study, we address these issues by modeling the decision-making process for carbon storage operations as a partially observable Markov decision process (POMDP). We solve the POMDP using belief state planning to optimize injector and monitoring well locations, with the goal of maximizing stored CO\textsubscript{2} while maintaining safety. Empirical results in simulation demonstrate that our approach is effective in ensuring safe long-term carbon storage operations. We showcase the flexibility of our approach by introducing three different monitoring strategies and examining their impact on decision quality. Additionally, we introduce a neural network surrogate model for the POMDP decision-making process to handle the complex dynamics of the multi-phase flow. We also investigate the effects of different fidelity levels of the surrogate model on decision qualities.
\end{abstract}

% keywords can be removed

\section*{Main}\label{sec1}
On our current trajectory, climate models project the global temperature will be $2.8^{\circ}$C hotter at the end of this century~\cite{UN2022report}, which almost doubles the $1.5^{\circ}$C target set by the Paris agreement. Failing to curb greenhouse emissions by transitioning to a low-carbon economy is catastrophic. A report published by the Intergovernmental Panel on Climate Change (IPCC) in 2018 suggests even a half a degree increase in global temperature from $1.5^{\circ}$C to $2^{\circ}$C can have significant negative impacts on human and natural systems~\cite{ipcc2018,franzke2014nonlinear}. As temperatures continue to rise, the implications and risks associated with climate change become more severe, affecting a wide range of areas from flood to drought, with disastrous consequences on food supply and ecosystems~\cite{parker2018fuel,trenberth2014global,norberg2012eco}. 

Many technologies can help to reduce global warming, such as renewable energy sources, technologies to improve efficiency, nuclear, and carbon capture and storage (CCS)~\cite{gernaat2021climate,abolhosseini2014review,siqueira2019current,wennersten2015future}. Among these technologies, CCS is the only one that can reduce $\mathrm{CO_2}$ emissions from large centralized sources~\cite{mitrovic2011carbon}. Many mitigation portfolios and emission reduction approaches, such as those assessed by IPCC and International Energy Agency (IEA), include CCS~\cite{ipcc2015,iea2020,peters2013challenge}. It is essential for addressing climate change and decarbonizing of the global energy system.

CCS is designed to capture $\mathrm{CO_2}$ emissions from large industrial sources, such as power plants and industrial facilities, and then transport and store them underground. The captured $\mathrm{CO_2}$ is typically injected into underground geological formations, such as depleted oil and gas reservoirs, saline aquifers, and coal seams~\cite{sci_report,commun}. Saline aquifers have the highest $\mathrm{CO_2}$ storage potential compared to other geological storage options, making them important prospects to scale CCS~\cite{michael2010geological}. 

By 2022, 20 countries are already developing and operating more than 130 $\mathrm{CO_2}$ storage sites~\cite{iea2022}. However, sequestering $\mathrm{CO_2}$ into geological formations for long-term storage comes with many subsurface technological challenges and safety concerns. Although $\mathrm{CO_2}$ sequestration projects may seem similar to other subsurface applications that involve working with the flow in porous media, such as groundwater or oil and gas~\cite{scheidt2018quantifying, tartakovsky2013assessment}, they pose unique challenges. First, less is known about saline aquifers than oil and gas reservoirs, resulting in significantly higher subsurface uncertainty. Second, one must account for the complex physical and chemical processes such as trapping mechanisms, mineral precipitations, and geomechanics that result from interactions between supercritical $\mathrm{CO_2}$, brine, and rock~\cite{wrr}. Injection of $\mathrm{CO_2}$ into geological formations comes with significant risks, such as induced earthquakes, fractured cap rocks, and reactivation of faults that can lead to $\mathrm{CO_2}$ leakage, harming human life and the environment~\cite{snaebjornsdottir2020carbon}. Thus, any decisions made in the course of a carbon storage project, such as injector locations, injection rates, and monitoring and mitigation strategies, are safety critical. Overall, the inherent subsurface uncertainty combined with complex physio-chemical processes imposes challenging decision problems for safe and effective carbon storage operations that require tight coupling of information gathering and sequential planning.

The Partial Observable Markov Decision Process (POMDP) is a framework for making sequential decisions under uncertainties (see Methods)~\cite{kochenderfer2022algorithms}. The framework has been successfully applied to many domains, such as autonomous driving, gaming, and healthcare~\cite{silver2017mastering,xiang2021recent,zhang2022diagnostic,hubmann2019pomdp}. Optimization of decision strategies within this framework has resulted in systems that perform better than human decision-makers and other state-of-the-art techniques~\cite{wang2022sequential}. 
\begin{figure}[htp]
     \centering
     \includegraphics[width=0.95\textwidth]{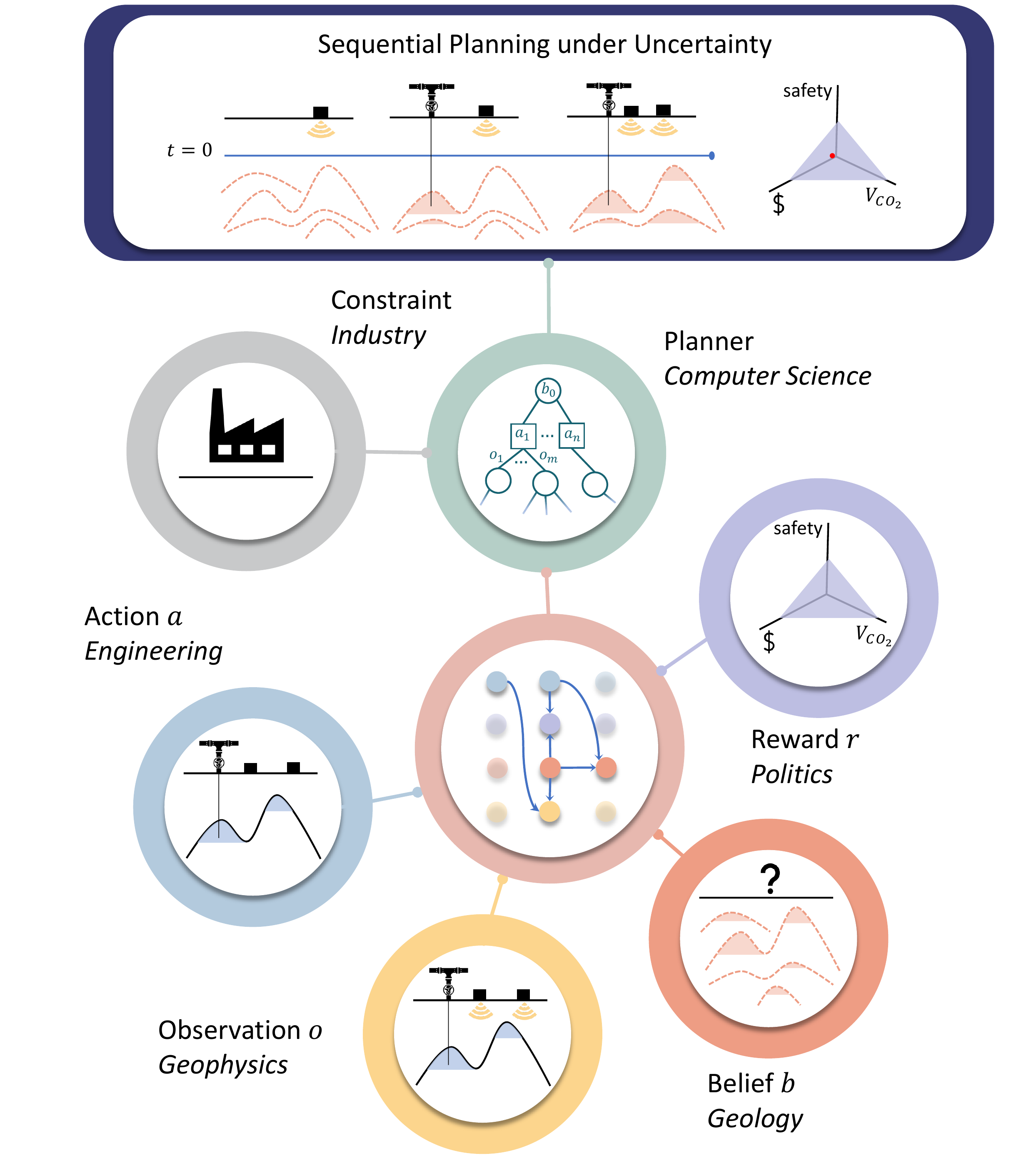}
     \caption{Optimizing for carbon storage operations.}
     \label{fig_general_cs}
\end{figure}

 As shown in Fig~\ref{fig_general_cs}, optimizing carbon storage operations requires multidisciplinary knowledge, including computer science, geology, engineering, economics, and politics. In a general POMDP formulation for carbon storage, the states may include the geological and engineering characteristics of a subsurface reservoir, flow properties, and joint space of well types, controls, and configurations. The actions may include running well tests, drilling wells, setting well controls, and even abandoning the project. The observations can come from various sources, including local high-resolution observations through boreholes and spatial observations such as seismic signals through active and passive monitoring~\cite{daley2007continuous,verdon2010passive}. The transition function is defined by the complex simulation of flow and transport in porous media that govern the injection and migration of the $\mathrm{CO_2}$ plume. The objective is to maximize the amount of sequestrated $\mathrm{CO_2}$ in a subsurface reservoir with some unknown properties and ensure safety for the long term. In carbon storage operations, it is necessary to consider the vast number of scenarios that arise from the high-dimensional state, action, and observation spaces over long planning horizons. This combinatorial problem presents a significant challenge for humans to find the optimal sequence of actions, especially when considering the incorporation of present and future information gathering. We propose an effective way to efficiently search for the best sequence of actions within this complex decision-making space that maximizes expected discounted return.

We formulate the carbon storage problem as a POMDP and solve it using approximate belief state planning. We show the importance of automated decision-making systems for ensuring safe operations by comparing algorithmically generated decisions against human judgments. We demonstrate how reasoning about future information is crucial for safety. In addition, we exemplify how planning over complex dynamics such as multi-phase flow becomes feasible when using a neural network surrogate model.

\section*{3D saline aquifer reservoir}
To demonstrate the usage of the decision systems in real-world carbon capture operations, we present a 3D saline aquifer reservoir. The aquifer is modeled by a set of engineering parameters and a 3D geological porosity map. We have $\mathrm{CO_2}$ injectors, which inject $\mathrm{CO_2}$ for a fixed period, and monitoring wells, which can be used to observe $\mathrm{CO_2}$ saturation. After the injection period, $\mathrm{CO_2}$ injection ceases. However, the $\mathrm{CO_2}$ still moves and spreads due to the gravitational and buoyancy forces for a more extended time, referred to as the post-injection period. The injected $\mathrm{CO_2}$ can be classified into three categories: exited, free, and trapped. The exited $\mathrm{CO_2}$ indicates the injected $\mathrm{CO_2}$ has leaked out of the aquifer. The free $\mathrm{CO_2}$ is $\mathrm{CO_2}$ that remains in the aquifer but is not physically bounded and is, therefore, movable. The trapped $\mathrm{CO_2}$ is already in equilibrium and chemically or physically trapped within the aquifer. 

\begin{figure*}[htp]
   \subfloat[]{\label{3d_res_bot}
    \hspace{0.6cm}
      \includegraphics[width=0.92\textwidth]{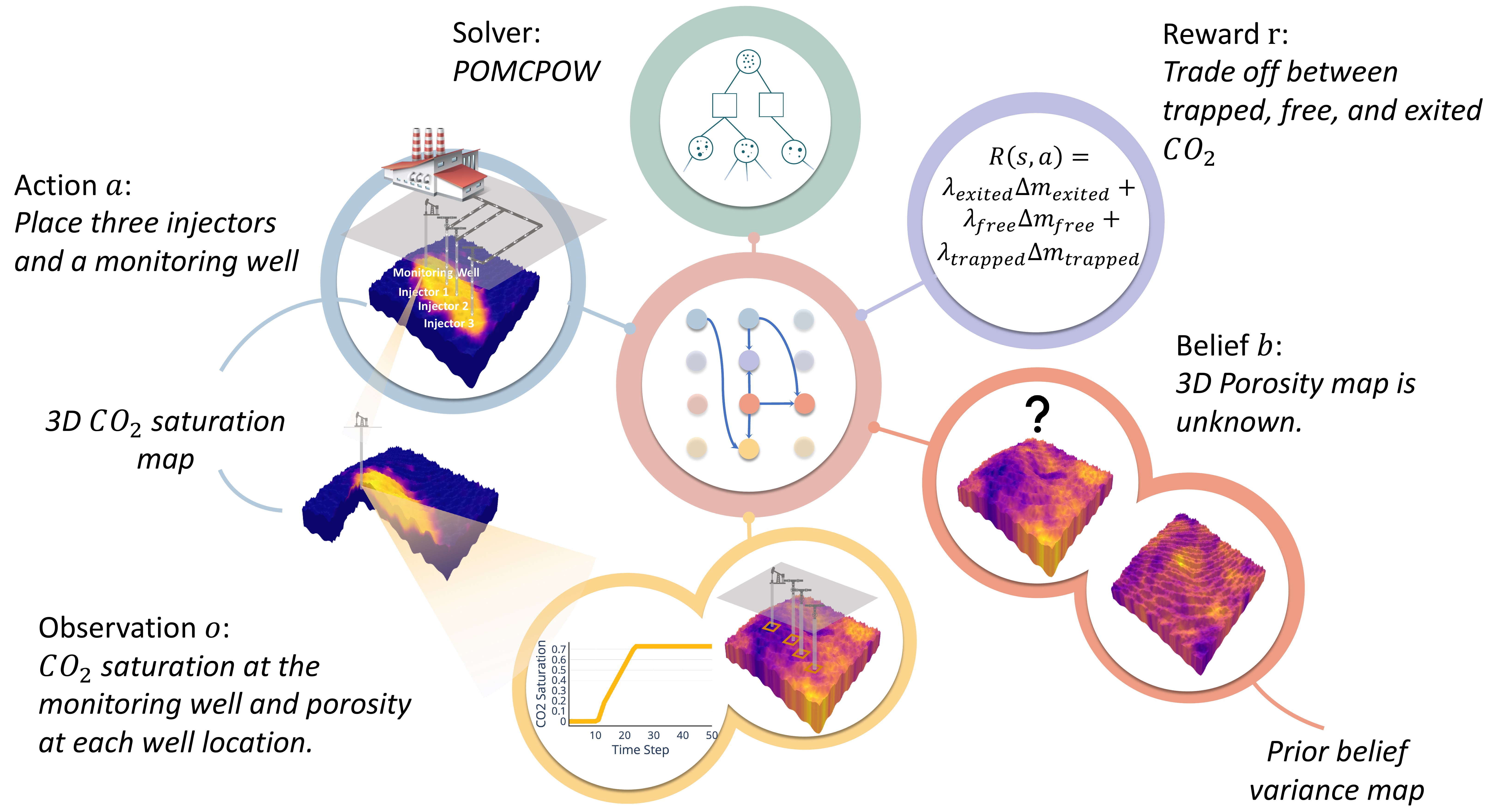}}
      
   \subfloat[]{\label{3d_res_top}
      \includegraphics[width=0.95\textwidth]{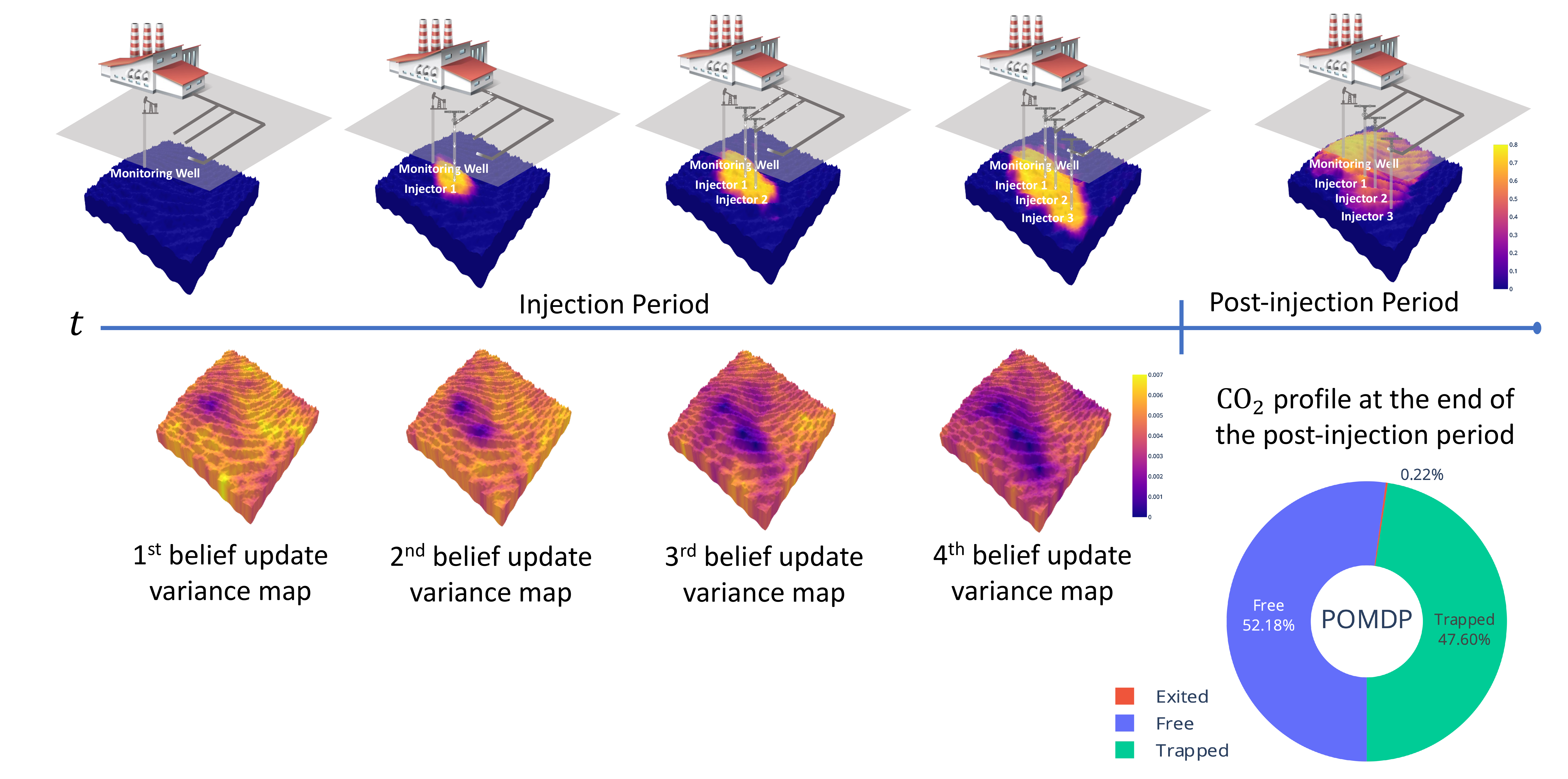}}

   \caption{3D saline aquifer carbon storage POMDP demonstration and its policy. (a) POMDP specification for individual components, emphasizing that the proposed belief state planning approach is a general methodology, without making specific claims about the superiority of a particular planner or parameter set. This allows operators and stakeholders to make trade-offs between various objectives. (b) The policy and overall performance of our approach in a single representative case, with 3D grids illustrating the aquifer structure.}\label{fig_3d_overall}
\end{figure*}

The POMDP formulation for the 3D example is shown in Fig.~\ref{3d_res_bot}. The state is defined by the joint space of well configurations, the amount of trapped, free, and exited $\mathrm{CO_2}$, and the 3D porosity map. The porosity field remains unknown. The action space is the locations where the wells may be placed, including both $\mathrm{CO_2}$ injectors and monitoring wells. All wells measure the porosity at the well locations. The monitoring wells observe a time history of $\mathrm{CO_2}$ saturation. All observations contain normally distributed noise. The transition dynamics are encoded in a multi-phase flow simulator. It simulates how $\mathrm{CO_2}$ migrates and calculates the mass of each portion of injected $\mathrm{CO_2}$ for the entire injection and post-injection periods. The reward function penalizes the total amount of $\mathrm{CO_2}$ escaping from the aquifer and the free $\mathrm{CO_2}$ and positively rewards the total trapped $\mathrm{CO_2}$ mass, where $\Delta m_{\rm exited}$, $\Delta m_{\rm free}$, and $\Delta m_{\rm trapped}$ are, respectively, the changes in exited, free, and trapped $\mathrm{CO_2}$ between states $s$ and $s^\prime$.
% \begin{equation}
% R(s, a, s^\prime) = \lambda_{\rm exited}\Delta V_{\rm exited} + \lambda_{\rm free}\Delta V_{\rm free} + \lambda_{\rm trapped}\Delta V_{\rm trapped} \label{eq:3d_reward}
% \end{equation}

\section*{Reducing computational time using a surrogate model}
While planning, the belief state planning approach simulates future scenarios and reasons how future information would affect the current decision. This process requires a large number of physical simulations. In our work, this implies our approach needs many simulation episodes of external multi-phase flow in porous media, which is highly computationally intensive (minutes to hours per simulation). To make planning feasible under a multi-phase flow environment, we use a deep learning-based surrogate model as an alternative to numerical simulation. Once trained, the surrogate model provides significant speed-up compared to traditional simulation while maintaining comparable accuracy. Therefore, surrogate models enable efficient repetitive forward simulations needed by our approach. 

\begin{figure}[htp]
    \centering
    \includegraphics[width=1.0\textwidth]{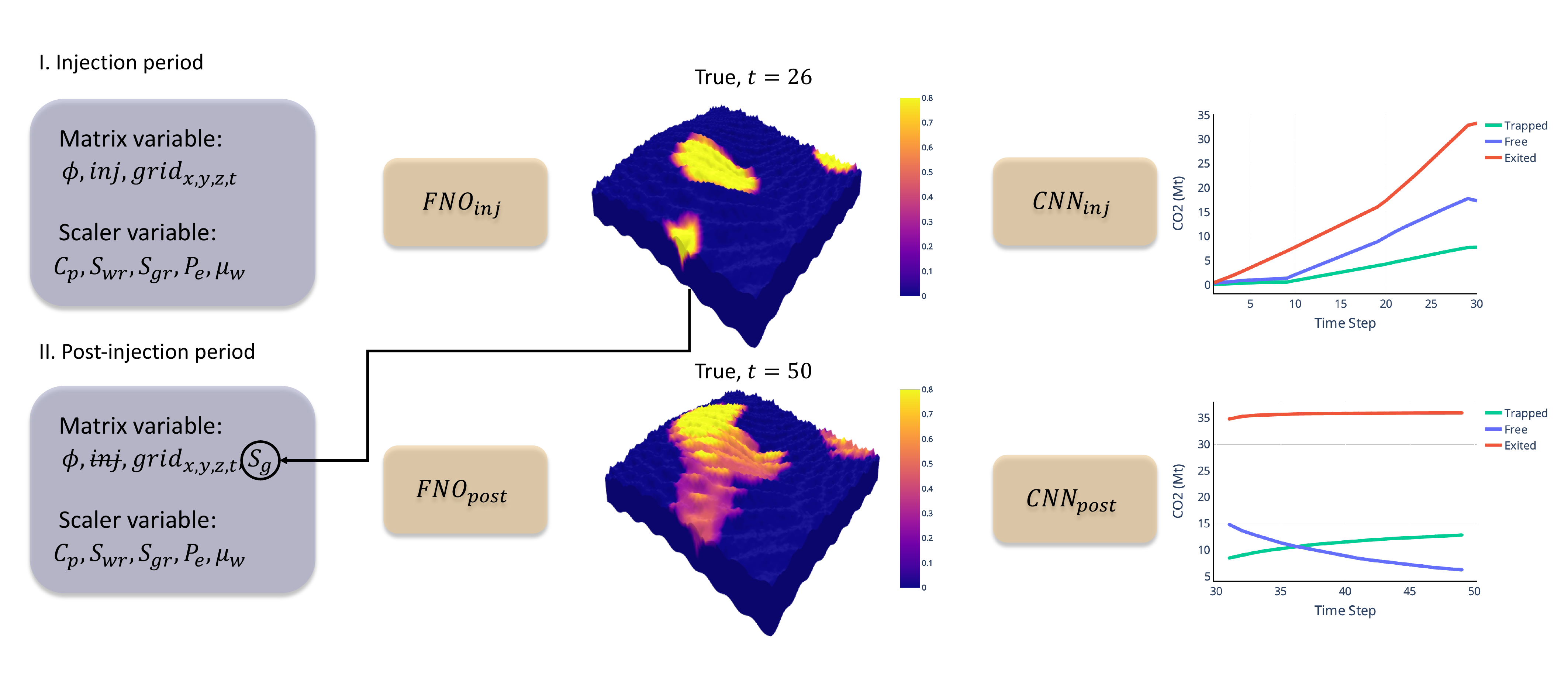}
    \caption{Surrogate model workflow for injection and post-injection period}
    \label{fig:fno}
\end{figure}

We use two surrogate models to predict multi-phase flow responses during the injection and post-injection period. As demonstrated in Fig.~\ref{fig:fno}, the surrogate model consists of two components: a Fourier Neural Operator (FNO)~\cite{li2020fourier} that predicts 4D (3D space-time) CO$_2$ gas saturation distribution, followed by a convolutional neural network (CNN) that predicts the temporal evolution of trapped, free, and exited CO$_2$ given 4D gas saturation distribution. 

FNO is a type of neural operator~\cite{li2020neural} that uses the Fourier integral kernel operator to learn empirical relationships given input-output data pairs. FNO-typed architecture has shown superior accuracy and data efficiency for CO$_2$ storage compared to the previous convolutional neural network-based surrogate model problems~\cite{wen2022u, wen2022accelerating}. The input data to the injection period FNO model includes a combination of well locations, drilling strategy, porosity heterogeneity, rock compressibility, irreducible water/gas saturation, capillary entry pressure, water viscosity, as well as spatial and temporal encoding. Since all injection wells are shut down during the post-injection period, the well location and drilling strategy information are not needed by the post-injection FNO model. Instead, we provide the injection period gas saturation map to aid the prediction of the post-injection $\mathrm{CO_2}$ migration.

We extend the original FNO architecture into 4D to predict the dynamic change of 3D gas saturation over 30 years of active injection and 500 years of post-injection monitoring. As shown in Fig.~\ref{fig:fno}, the 3D spatial domain of the training data is an irregular geological formation. This leads to complex interactions between the flow behavior and gravity. The onset of and location of each injection well further increases the dimensionality of this problem. Nevertheless, the FNO architecture successfully learns this complex input-output 4D mapping with minimal overfitting. After obtaining the gas saturation prediction, we then use a CNN model to map the dynamic 3D gas saturation at each time step into time series predictions of $\mathrm{CO_2}$ mass. The CNN architecture consists of three channels for trapped, free, and mobile $\mathrm{CO_2}$ masses (see Extended Data Table~\ref{model_1}--\ref{model_4} for details of the surrogate architecture and parameters). All training and test datasets utilized in this work were generated employing the MATLAB Reservoir Simulation Toolbox (MRST) ~\cite{lie2019introduction}, a versatile and widely recognized computational tool in the domain of reservoir simulations. Using the surrogate model, we obtain predictions with three orders of magnitude speed-up compared to traditional numerical simulators.

\section*{Results for the 3D case}
In this work, we compare our proposed method with two baseline approaches. The first baseline, referred to as the random policy, involves selecting injectors randomly. The second baseline, called the expert policy, relies on experts utilizing their knowledge of the aquifer's geological characteristics and the subsurface flow processes to determine injector locations. We evaluate the performance of the random, expert, and belief state planning policies.

\begin{figure*}[htp]
   \subfloat[]{\label{3d_single_a}
    \hspace{0.6cm}
      \includegraphics[width=0.85\textwidth]{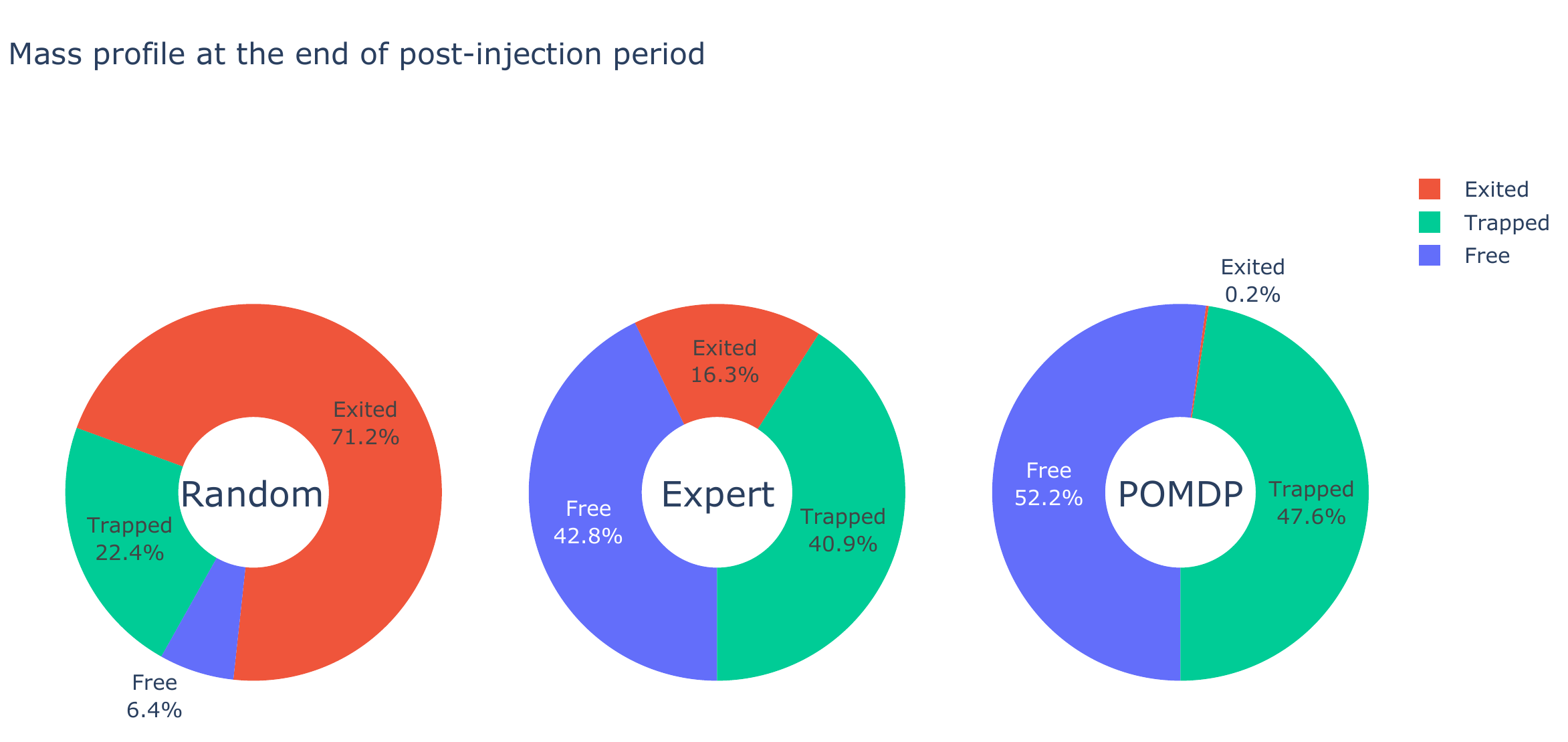}}
      
   \subfloat[]{\label{3d_agg_b}
      \includegraphics[width=0.95\textwidth]{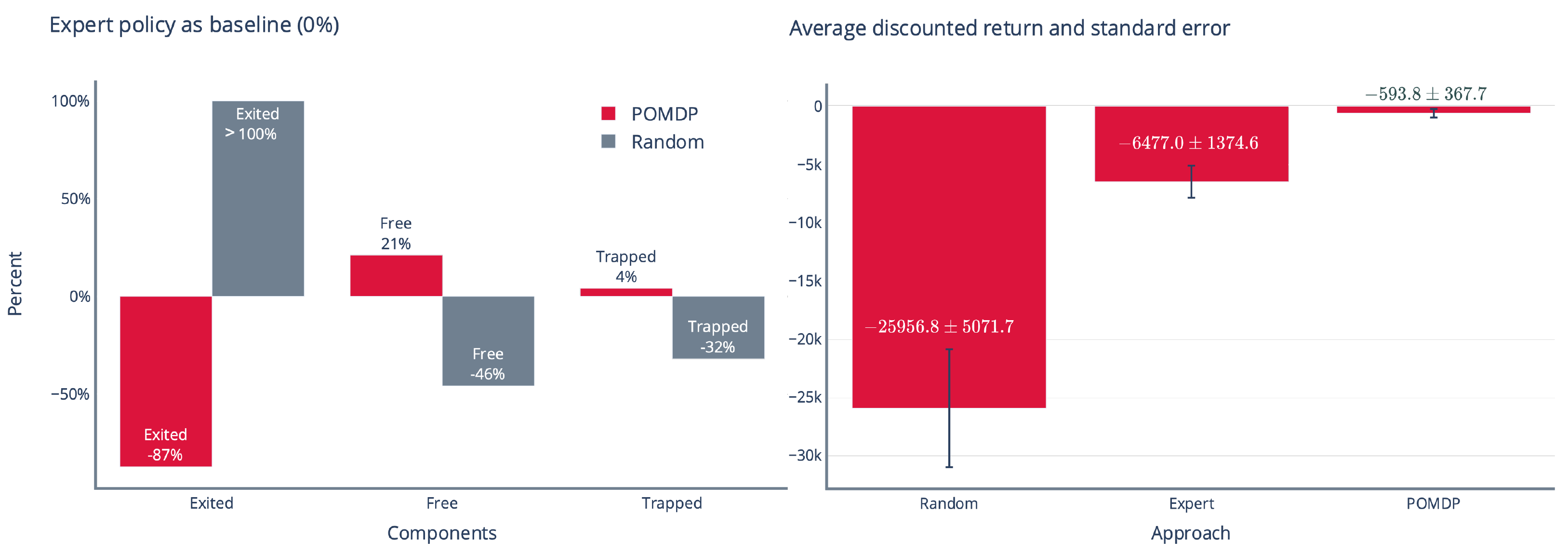}}

   \caption{Performance comparison of three solution methods on the 3D example. (a) $\mathrm{CO_2}$ mass profiles at the end of the post-injection period for a single representative case.   (b) Aggregated performance metrics across ten cases.}\label{fig_2d}
\end{figure*}

Our approach makes decisions based on its belief at each time step. To illustrate the belief regarding the true porosity map, we present the variance maps of 100 sampled porosity maps from the belief after each decision step (posterior) in Fig.~\ref{3d_res_top}. Darker colored areas signify lower variance, as the data acquired through drilling a well and observing $\mathrm{CO_2}$ saturation over time via the monitoring well lead to greater certainty in the belief. Fig.\ref{3d_single_a} summarizes the results due to different approaches on a single representative case. Our approach demonstrates superior performance, resulting in a 16.3\% increase in trapped $\mathrm{CO_2}$ and a 98.8\% reduction in leaked $\mathrm{CO_2}$ compared to the expert policy at the end of the post-injection period. 

To comprehensively evaluate and compare the three approaches, we test their policies on ten cases with different 3D porosity maps as ground truths. Compared to the expert policies, the policies that result from our approach end up achieving 87\% less exited $\mathrm{CO_2}$ and 4\% more trapped $\mathrm{CO_2}$ (see~Fig.\ref{3d_agg_b}). On the other hand, the random policies have a 2.9-fold increase in the quantity of exited $\mathrm{CO_2}$ when compared to the expert policies. In addition, we use both mean and standard error of discounted return as metrics. As a result, our approach attains the highest expected discounted return across ten cases. A larger expected discounted return suggests a safer and more effective carbon capture policy. A smaller standard error of the discount return indicates a more robust performance around the expected discounted return. The results suggest our approach can provide robust performance compared to the other two baseline methods, ensuring safe and effective carbon storage operations.

\section*{Effects of Surrogate Model Fidelity Levels on Decision Quality}
In this work, we use a trained surrogate model as a proxy for the complex decision environment, raising the question of how planning with different fidelity levels of the surrogate model might affect the performance of our proposed approach. To address this question, we trained surrogate models at two distinct fidelity levels. The high-fidelity model exhibits significantly smaller root mean squared errors (RMSE) in predicting $\mathrm{CO_2}$ mass over time compared to its low-fidelity counterpart. For instance, the RMSE of trapped, free, and exited $\mathrm{CO_2}$ mass predictions for the high-fidelity model are 1.1 MT, 1.8 MT, and 2.3 MT, respectively. Conversely, the corresponding RMSE values for the low-fidelity model are 1.8 MT, 2.8 MT, and 3.6 MT.

\begin{figure*}[htp]
   \centering
   \subfloat[]{\label{fid_a}
      \includegraphics[width=0.49\textwidth]{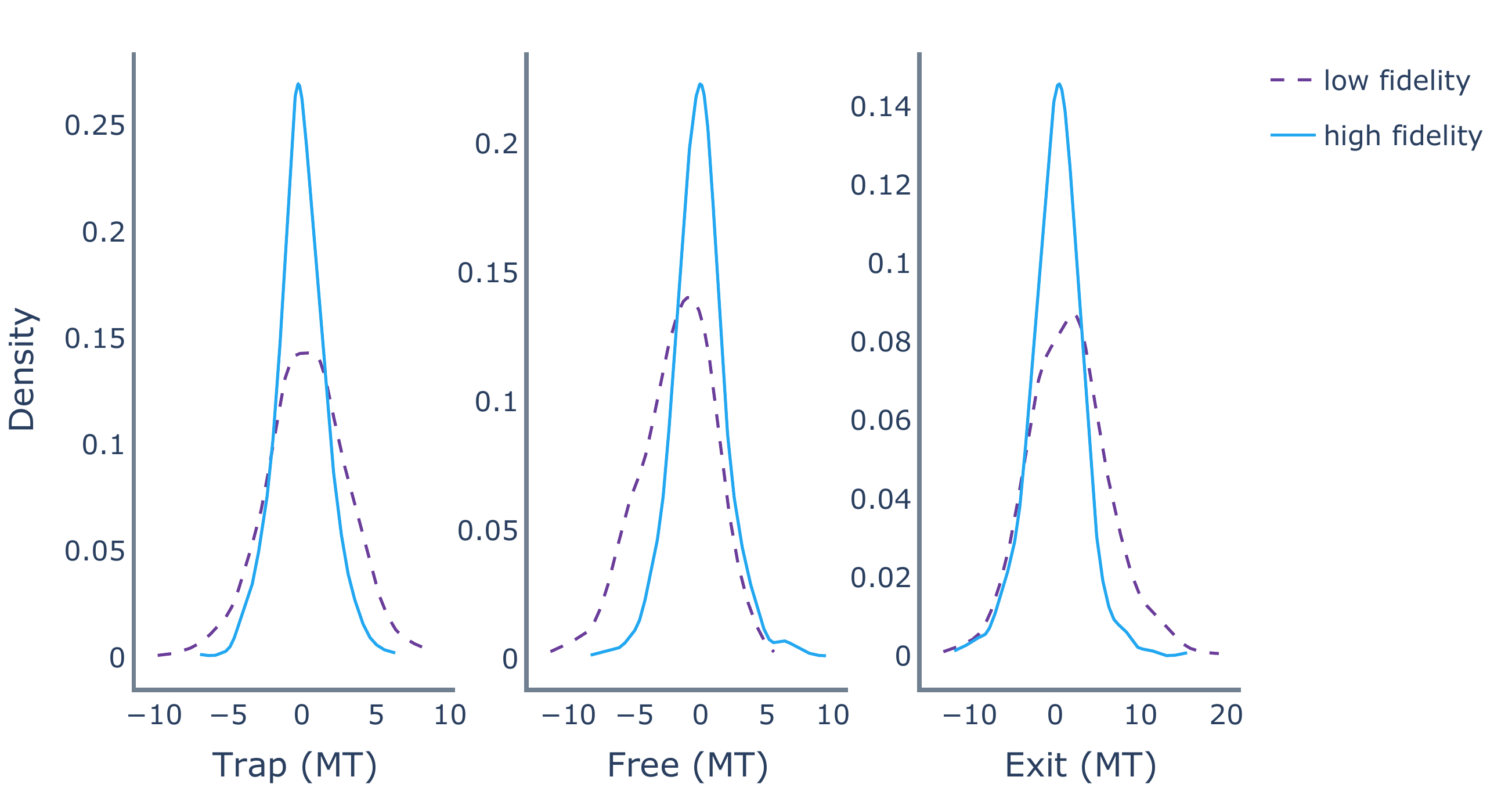}}
   \hfill
   \subfloat[]{\label{fid_b}
      \includegraphics[width=0.49\textwidth]{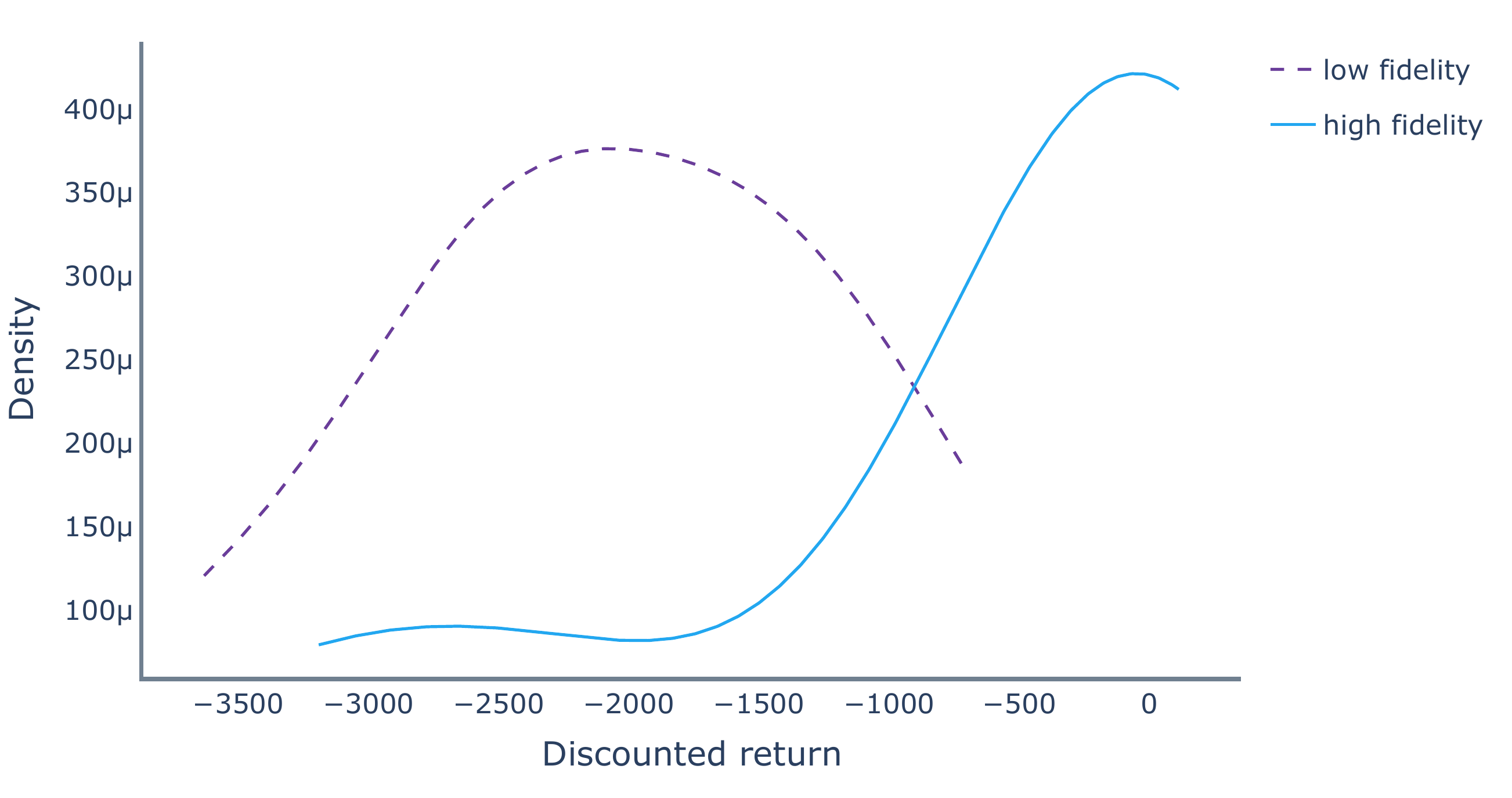}}
   \caption{Comparative analysis of surrogate model fidelity. (a) Density plots illustrating the errors between the high and low fidelity surrogate predictions and MRST outputs. (b) Density plot on discounted return for high and low fidelity surrogate models across ten cases.}\label{fig_2d}
\end{figure*}

When comparing the density plots of errors between high and low fidelity surrogate model predictions of $\mathrm{CO_2}$ mass and MRST outputs at the end of the post-injection period, we find that the error density plots of the high fidelity model are more closely centered around 0 and exhibit smaller variance than those of the low fidelity model (see Fig.~\ref{fid_a}). We calculated the JS-divergences between the density curves, which are 0.0075, 0.0126, and 0.0057, respectively. The fidelity level of the surrogate models used in planning has a significant impact on decision quality. Fig.~\ref{fid_b} shows that when planning with the high fidelity model, the density curve of the discounted return moves significantly to the right, indicating safer and more effective policies than planning with the low fidelity model. The JS-divergence between the two density plots on the discounted return due to high and low fidelity models is 42.2549, highlighting the significant impact that even small changes in the fidelity of the surrogate model can have on overall discounted returns. 

In light of these results, the question naturally arises:``How accurate is accurate enough?" While it is clear that model accuracy plays a crucial role in the quality of the decision-making process, it is challenging to determine an exact threshold for sufficient accuracy. Practically, perfect accuracy is often unattainable, and operators must strike a balance between computational cost and model fidelity. The key takeaway from our analysis is that accuracy matters, but the surrogate model does not have to be perfect. We recommend that, as a best practice, any plan derived from a surrogate model should be validated using a higher fidelity model before implementation. This ensures that the most critical decisions are based on the most accurate information available, while still benefiting from the computational efficiency provided by surrogate models during the planning process.

\section*{Demonstrating the value of the monitoring well}
As shown in Fig.~\ref{fig_3d_overall}, our approach needs to make sequential decisions on the locations of both the monitoring well and injectors. While it is common practice to include a monitoring well in the field to obtain subsurface information and monitor $\mathrm{CO_2}$ saturation, our study, for the first time, quantifies the utility of the monitoring well in this context. Adding the monitoring well increases the complexity of making sequential decisions because of the alignment of the monitoring well with injectors to maximize information gain and better inform future actions. To evaluate the value and necessity of the monitoring well, we use the same POMDP formulation except without the monitoring well in the action space, resulting in a different observation space. 

\begin{table}[h]
\begin{center}
\begin{minipage}{\textwidth}
\caption{Our proposed approach performance comparison under different observation strategies}\label{moni}
\begin{tabular*}{\textwidth}{@{\extracolsep{\fill}}lllll@{\extracolsep{\fill}}}
\toprule%
\multirow{3}{*}{Plan scenarios} & \multicolumn{1}{l}{\multirow{2}{*}{Discounted return}}   & \multicolumn{3}{@{}c@{}}{$\mathrm{CO_2}$ mass profile (MT)$^{a}$} \\\cmidrule{3-5}%
                                &  \multicolumn{1}{l}{(mean$\pm$SE)}                            & \multicolumn{1}{l}{Trapped}  & \multicolumn{1}{l}{Free}  & \multicolumn{1}{l}{Exited} \\
\midrule
No monitoring well & $-1262.9 \pm 485.3$                         & 30.5         & 26.9              & 1.6  \\
Monitoring well    & $-593.8 \pm 367.7$                          & $\mathbf{30.9}$         & 27.2              & 0.9   \\
4D seismic        & $\mathbf{-475.8} \pm \mathbf{157.8}$        & 30.4         & 27.8              & $\mathbf{0.8}$   \\
\bottomrule
\end{tabular*}
\footnotetext[1]{The values presented in the $\mathrm{CO_2}$ mass profile represent the data collected at the conclusion of the post-injection phase.}
\end{minipage}
\end{center}
\end{table}

As shown in the first two rows of Table~\ref{moni}, planning with the monitoring well has a higher expected discounted reward and a lower standard error across ten cases. This suggests that the extra information from the monitoring well can enhance the decision quality. 

\section*{Extending the point-wise observations to spatial observations}
One of the key challenges in safe carbon storage operations is the lack of knowledge on $\mathrm{CO_2}$ plume migration under complex geological regimes while planning. Geophysical monitoring (e.g., through seismic surveys) can provide predictions on $\mathrm{CO_2}$ plume saturation and delineate its boundaries~\cite{fawad2021monitoring}. To demonstrate the possibility of adding seismic observations to the carbon storage POMDP, we replace placing the monitoring well with the ability to make seismic observations before drilling each injector in the action space of the 3D example. We mimic the seismic saturation observations by adding spatial filtering with Gaussian kernel~\cite{bhabatosh1977digital} over the underlying true saturation map (see Extended Data Fig.~\ref{fig_blur} for an example). Among all the observation strategies outlined in Table~\ref{moni}, employing seismic observations in the planning process results in the safest policies and yields the highest discounted return and the lowest standard errors associated with it.

\section*{Discussion and conclusion}

In this study, we address a sequential decision problem in carbon storage by employing belief state planning to ensure safe and effective operations. Our method significantly outperforms expert decisions in terms of expected discounted return and standard error across ten cases. For instance, in the representative 3D case, our proposed approach reduces the amount of leaked $\mathrm{CO_2}$ by over 87\% when compared to the expert policy. Furthermore, we find that incorporating information from monitoring wells and seismic observations allows our approach to achieving safer and more effective policies.

To make planning under a multi-phase flow environment feasible, we train an FNO neural network as a surrogate flow simulator. We investigate the impact of different fidelity levels of the surrogate model on the decision quality of our proposed approach. Our results indicate that planning with a high-fidelity model leads to substantially safer and more effective decisions for carbon storage operations. Large-scale simulations serve as an effective means of evaluating our methodology, as physical experiments would not only take years to complete but also fail to adequately assess the robustness of the decision strategies. Moreover, it would be challenging to fairly compare various methodologies across a broad range of scenarios using physical experiments alone.

As the development of carbon storage projects gains momentum globally, ensuring the safe operation of these initiatives is of paramount importance. Our research highlights the potential necessity of our proposed approach in ensuring the long-term safety of carbon storage projects.

\section*{Data availability}
The data used for this work is made available in the repository https://github.com/yizhengw/OptimizingCarbonStorage.

\section*{Code availability}
The code used to produce the results for this work is made available in the repository https://github.com/yizhengw/OptimizingCarbonStorage.

\bibliographystyle{unsrt}  
\bibliography{references}

\section*{Methods}\label{method}

\subsection*{POMDPs}
\label{pomdp_review}
A POMDP is a model for sequential decision making under uncertainty. A POMDP is defined as a tuple ($\mathcal{S}, \mathcal{A}, \mathcal{O}, O, T, R, \gamma$), where $\mathcal{S}$ is the set of states, $\mathcal{A}$ is the set of actions, $\mathcal{O}$ is the set of observations, $O$ is the observation function, $T$ is the state transition function, $R$ is the reward function, and $\gamma$ is the discount factor. At each step, an agent in state $s \in S$ takes action $a \in \mathcal{A}$, transitions to the next state $s' \in \mathcal{S}$ with probability $T(s'\mid s, a)$ and receives both an observation $o \in \mathcal{O}$ with probability $O(o \mid s', a)$ and a reward $r = R(s, a, s')$. In a POMDP, the underlying state of the environment is not fully known. Instead, the agent maintains a distribution over states called a belief $b(s)$. The agent infers its belief from the history of observations and actions according to Bayes's rule, $b'(s')=p(s'\mid b, a, o) \propto O(o \mid s', a)\sum_{s \in \mathcal{S}}T(s' \mid s, a)b(s)$. 

A policy is a function that maps beliefs to actions. The goal is to find the optimal policy $\pi^*$ that maximizes the expected sum of discounted reward, $\sum_{t=1}^n \gamma^{n-1}r_t$.  For the infinite horizon sequential problem where $n \to \infty$, the discounted factor $\gamma \in [0,1)$ is introduced to bound the total discounted reward.

In general, there are two approaches to generating a policy for POMDPs: offline or online during action execution. Offline solvers perform most of the computation prior to interacting with the environment while online solvers perform computation between state transitions.

\subsection*{3D saline aquifer example}
\label{3d_setup}
The 3D aquifer grids have an extension of eighty grid blocks in the x- and the y-direction (lateral extension) and eight blocks in the z-direction (depth). There are three injectors and one monitoring well. The monitoring well needs to be placed before any injectors. The three injectors are placed in a time sequence at the 1st, 11th, and 21st years. The injection period and post-injection period, respectively, are 30 and 500 years resulting in a total of 530 years of simulation time. We assume permeability $K$ follows a deterministic function of porosity $\phi$, with $K=\frac{10^{-10}\phi^3}{58.32(1-\phi)^2}$. Therefore, our approach only needs to keep track of the porosity map in its state space and observes porosity values at each well location (see Extended Data Table~\ref{tab:parameters}-\ref{tab:hyperparameters} for specific values of geological, engineering, POMDP, and planner parameters).

\subsection*{Surrogate model training details} 
\label{suro_train}
For both the injection and post-injection surrogate models, we train the saturation and mass separately. We trained the gas situation FNO models using 300 training samples and 100 test samples for around 300 epochs. Each epoch takes approximately 3 mins. For the $\mathrm{CO_2}$ mass CNN models, we trained with 600 training samples and 100 test samples for around 300 epochs. Each epoch takes approximately 11 s. $\mathrm{CO_2}$ mass requires more training samples because it is more prone to overfitting. The relative squared-error loss is used for both scenarios to ensure effective gradient propagation. After training, the average gas saturation error given unseen inputs is 0.013 during the injection period and 0.016 during the post-injection period. The average mass error (defined as the mean absolute error divided by the mass norm) is 0.012 during the injection period and 0.015 during the post-injection period. To obtain surrogate models with two distinct fidelity levels, we utilize 12,000 training data for the high-fidelity model and 1,700 training data for the low-fidelity model. 

\subsection*{Partially Observable Monte Carlo Planning with Observation Widening} 
\label{pomcpow_review}
Due to the relatively long time scales expected in CCS, we use an online solver called Partially Observable Monte Carlo Planning with Observation Widening (POMCPOW). POMCPOW is a tree search algorithm used for online decision-making, which can effectively handle continuous state, action, and observation spaces~\citesupp{sunberg2018online}. Prior to POMCPOW, state-of-the-art online planers such as POMCP and DESPOT struggled to solve POMDPs with continuous observation spaces~\citesupp{somani2013despot,silver2010monte}. Their planning trees suffer from belief collapse beyond the root node and end up generating suboptimal policies. POMCPOW, on the other hand, creates weighted particle-based search trees, resulting in richer belief representations at each node, allowing the search trees to grow deeper.
Richer beliefs also make generated policies contain information-gathering actions and thus yield better strategies for the problems where information-gathering is needed.

\subsection*{Ensemble Smoother with Multiple Data Assimilation}
\label{esmda_review}
Ensemble Smoother with Multiple Data Assimilation (ES-MDA) is a commonly used ensemble-based inference method for estimating state uncertainties~\citesupp{emerick2013ensemble}. Unlike Bayesian inversion using Markov chain Monte Carlo (MCMC) methods to obtain posterior distributions, ensemble-based methods can significantly reduce the computational cost of belief updates. Among the ensemble-based methods, the ensemble Kalman filter (EnKF) is one of the most popular approaches~\citesupp{evensen2003ensemble}. However, the recurrent simulation restarts required in the EnKF sequential data simulation process make this method impractical for problems that involve expensive forward simulations. On the other hand, in ensemble smoother (ES), all data are assimilated simultaneously in a single update~\citesupp{skjervheim2011ensemble}. Therefore, there is no need for simulation restarts, which makes it more attractive for practical applications. However, because ES computes a single update, it may not yield a good result that matches the data. In order to fix this problem, ES-MDA introduces an inflation factor ($\alpha$) in front of the data-error covariance matrix $C_D$. This factor helps the algorithm to correctly sample from the posterior distribution~\citesupp{emerick2016analysis}.

\subsection*{Belief update}
The porosity map is partially observable. However, the belief over the porosity map can be informed by both direct measurements of aquifer porosity at each well location and $\mathrm{CO_2}$ saturation history. Because observations are obtained from two different sources, the belief update is formulated as follows:
\begin{equation}
p(\mathbf{\phi} \mid d_1, \mathbf{d_2}) = \sum_{\mathbf{\phi_{cond}}}p(\mathbf{\phi} \mid \mathbf{\phi_{cond}}, \mathbf{d_2})p(\mathbf{\phi_{cond}}\mid d_1)
\label{belief_3d}
\end{equation}
where $\mathbf{\phi}$ stands for the porosity maps, $d_1$ is the observed porosity value, and $\mathbf{d_2}$ is the observed $\mathrm{CO_2}$ saturation history at the monitoring well. The term $\mathbf{\phi_{cond}}$ represents the conditional porosity maps with only $d_1$ using conditional Gaussian simulation~\citesupp{hoshiya1995kriging}. Therefore, we can assume $\mathbf{\phi}$ is conditional independent with $d_1$, statistically represented as $p(\mathbf{\phi} \mid d_1, \mathbf{d_2}, \mathbf{\phi_{cond}})=p(\mathbf{\phi} \mid \mathbf{d_2}, \mathbf{\phi_{cond}})$.

To reduce the computational cost of solving the inversion problem with high dimensional data, we deploy ES-MDA to sample from the posterior of $\mathbf{\phi}$ given both $\mathbf{d_2}$ and $\mathbf{\phi_{cond}}$.

\bibliographystylesupp{unsrt}
\bibliographysupp{references}

\renewcommand{\figurename}{Extended Data Fig.}
\setcounter{figure}{0}

\begin{figure}[htp]
    \centering
    \includegraphics[width=0.95\textwidth]{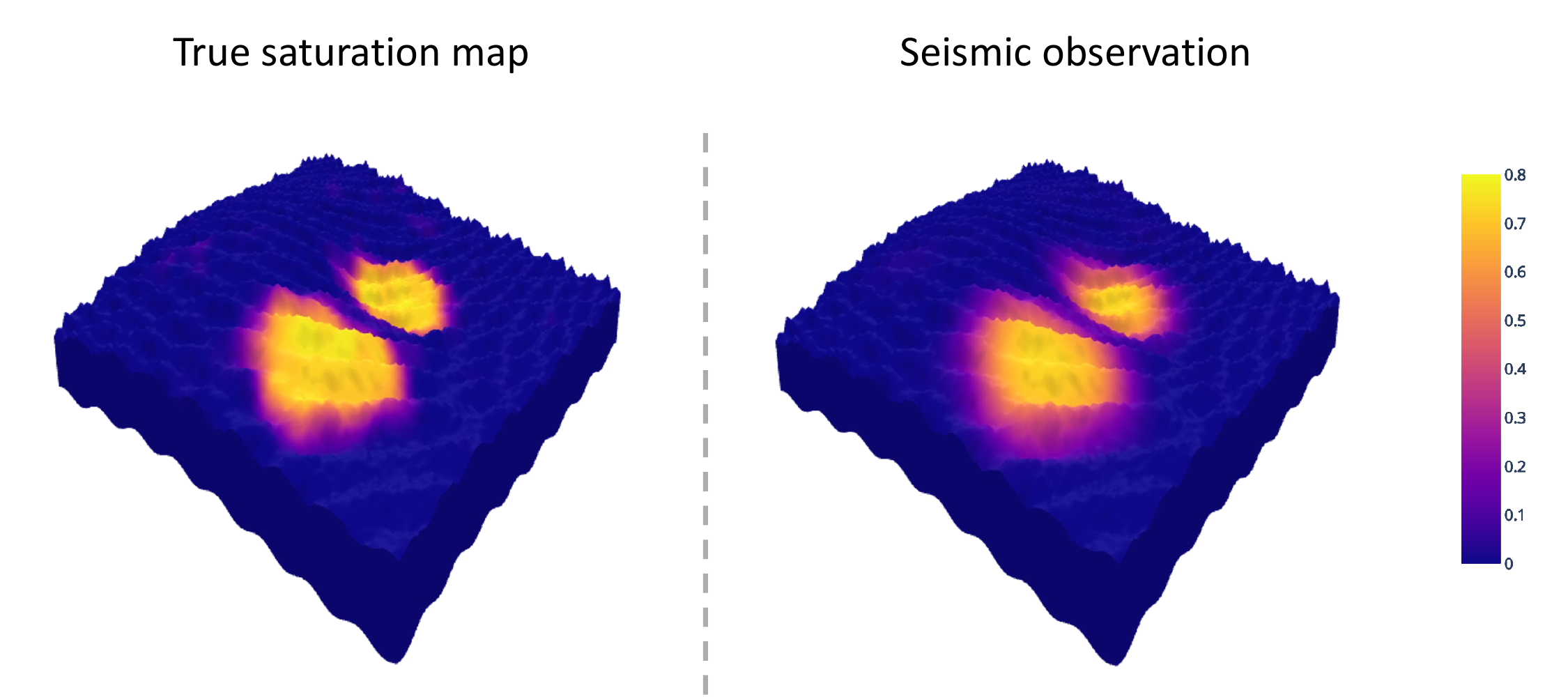}
    \caption{Seismic observation example for a given 3D saturation map}
    \label{fig_blur}
\end{figure}

\renewcommand{\tablename}{Extended Data Table}
\setcounter{table}{0}

% \begin{table}[h]
% \begin{center}
% \begin{minipage}{\textwidth}
% \caption{Comparison of 2D carbon storage policies}\label{2d_res}%
% \centering
% \begin{tabular}{@{}lr@{\ \pm \ }lrrr@{}}
% \toprule
% \multirow{2}{*}{Method}  &   \multicolumn{2}{c}{Discounted Return}      &  Leak Fraction      & \multicolumn{2}{c}{Trap Efficiency} \\
%                          &  \multicolumn{2}{c}{(mean $\pm$ SE)}         &                     & \multicolumn{2}{c}{(mean $\pm$ SE)}\\
% \midrule
% Random Policy           & $-104.38        &320.41$                       & $0.10$            & $0.08  &0.07$ \\
% Expert Policy           & $-97.91         &316.28$                       & $0.10$            & $\mathbf{0.82} & \mathbf{0.17} \\
% POMCPOW                 & $\mathbf{2.24}  &\mathbf{4.04}$                & $\mathbf{0.00}$   & $0.27 & 0.32   \\
% \botrule
% \end{tabular}
% \end{minipage}
% \end{center}
% \end{table}

\begin{table}[h]
\begin{center}
\begin{minipage}{\textwidth}
    \caption{Geological and engineering parameters for the 3D aquifer\label{tab:parameters}}
    \centering
    \begin{tabular}{@{}ll@{}} 
    \toprule
    Variable Description & Value \\
    \midrule
    Porosity map mean & 0.2\\
    Porosity map variogram sill & 0.001\\
    Porosity map variogram range & 20.0\\
    Porosity map variogram nugget & 0.0001\\
    \midrule
    Rock compressibility & $\num{4.35e-5}$\\
    Irreducible water saturation & 0.27 \\
    Irreducible gas saturation & 0.20 \\
    Capillary entry pressure & 5.0\\
    Water viscosity& $\num{8e-4}$\\
    \bottomrule
    \end{tabular}
\end{minipage}
\end{center}
\end{table}

\begin{table}[h]
\begin{center}
\begin{minipage}{\textwidth}
    \caption{POMDP and POMCPOW hyperparameters\label{tab:hyperparameters}}
    \centering
    \begin{tabular}{@{}ll@{}} 
    \toprule
    Variable Description & Value \\
    \midrule
    Discount Factor ($\gamma$) & 0.99 \\
    $\lambda_{\rm exited}$ & -1000.0\\
    $\lambda_{\rm free}$ & -1.0\\
    $\lambda_{\rm trapped}$ & 10.0\\
    Gaussian filtering kernel $\sigma$ & 3.0 \\
    Distribution of observation noise & $\mathcal{N}(0, 0.1\times\phi_{obs})$\\
    \midrule
    Exploration Coefficient for UCB & 20.0 \\
    Observation widening exponent ($\alpha_{\rm obs})$ & 0.7 \\
    Observation widening coefficient ($k_{\rm obs})$ & 5.0 \\
    Action widening exponent ($\alpha_{\rm act}$) & 0.7 \\
    Action widening coefficient ($k_{\rm act})$ & 2.0 \\
    Number of tree queries ($N_{\rm query}$) & 100\\
    \bottomrule
    \end{tabular}
\end{minipage}
\end{center}
\end{table}

\begin{table}[h]
\begin{center}
\begin{minipage}{\textwidth}
  \centering
  \caption{Model parameters for gas saturation during the injection period. The \texttt{Padding} denotes a padding operator that accommodates the non-periodic boundaries; \texttt{Linear} denotes the linear transformation to lift the input to the high dimensional space and the projection back to the original space; \texttt{Fourier4d} denotes the 4D Fourier operator; \texttt{Conv1d} denotes the bias term; \texttt{Add} operation adds the outputs together; \texttt{GELU} denotes a Gaussian Error Linear Units layer. }\label{model_1}
  \footnotesize
  \begin{tabular}{@{}lll@{}}
    \toprule
    Layer     & Operation     & Output Shape \\
    \midrule
    Input    & -                                    & (30, 80, 80, 8, 11) \\
    Padding &   \texttt{Padding (4)}                & (38, 88, 88, 16, 11) \\
    Lifting &  \texttt{Linear}                      & (38, 88, 88, 16, 28)  \\
    Fourier 1 & \texttt{Fourier4d/Conv1d/Add/GELU}  & (38, 88, 88, 16, 28) \\
    Fourier 2 & \texttt{Fourier4d/Conv1d/Add/GELU}  & (38, 88, 88, 16, 28) \\
    Fourier 3 & \texttt{Fourier4d/Conv1d/Add/GELU}  &(38, 88, 88, 16, 28)\\
    Fourier 4 & \texttt{Fourier4d/Conv1d/Add}  & (30, 80, 80, 8, 28)  \\
    De-padding     &  \texttt{Depadding (4)}                    & (30, 80, 80, 8, 28) \\
    Projection 1   &  \texttt{Linear}                          & (30, 80, 80, 8, 64) \\
    Projection 2   &  \texttt{Linear}                          & (30, 80, 80, 8, 1) \\
    \bottomrule
  \end{tabular}
  \label{table:injfno}
  \end{minipage}
\end{center}
\end{table}

\begin{table}[h]
\begin{center}
\begin{minipage}{\textwidth}
  \centering
  \caption{Model parameters for CO$_2$ mass during the injection period. After the flattening layer, the model splits into three channels where each channel contains an FC1 and FC2.}\label{model_2}
  \footnotesize
  \begin{tabular}{@{}lll@{}}
    \toprule
    Layer     & Operation     & Output Shape \\
    \midrule
    Input & -                         & (30, 80, 80, 8) \\
    Conv1 & \texttt{Conv3D/BN/ReLU}   & (30, 40, 40, 4) \\
    Conv2 & \texttt{Conv3D/BN/ReLU}   & (30, 20, 20, 2)  \\
    Flatten & - & (1, 24000) \\
    FC1 & \texttt{FC/BN/ReLU}   & (1, 100) \\
    FC2 & \texttt{FC/BN/ReLU}   & (1, 30) \\
    \bottomrule
  \end{tabular}
  \label{table:injmass}
    \end{minipage}
\end{center}
\end{table}

\begin{table}[h]
\begin{center}
\begin{minipage}{\textwidth}
  \centering
  \caption{Model parameters for gas saturation during post-injection period. The \texttt{Padding} denotes a padding operator that accommodates the non-periodic boundaries; \texttt{Linear} denotes the linear transformation to lift the input to the high dimensional space and the projection back to the original space; \texttt{Fourier4d} denotes the 4D Fourier operator; \texttt{Conv1d} denotes the bias term; \texttt{Add} operation adds the outputs together; \texttt{GELU} denotes a Gaussian Error Linear Units layer. }\label{model_3}
  \footnotesize
  \begin{tabular}{@{}lll@{}}
    \toprule
    Layer     & Operation     & Output Shape \\
    \midrule
    Input    & -                                    & (20, 80, 80, 8, 11) \\
    Padding &   \texttt{Padding (4)}                & (28, 88, 88, 16, 11) \\
    Lifting &  \texttt{Linear}                      & (28, 88, 88, 16, 28)  \\
    Fourier 1 & \texttt{Fourier4d/Conv1d/Add/GELU}  & (28, 88, 88, 16, 28) \\
    Fourier 2 & \texttt{Fourier4d/Conv1d/Add/GELU}  & (28, 88, 88, 16, 28) \\
    Fourier 3 & \texttt{Fourier4d/Conv1d/Add/GELU}  &(28, 88, 88, 16, 28)\\
    Fourier 4 & \texttt{Fourier4d/Conv1d/Add}       & (20, 80, 80, 8, 28)  \\
    De-padding     &  \texttt{Depadding (4)}                    & (20, 80, 80, 8, 28) \\
    Projection 1   &  \texttt{Linear}                          & (20, 80, 80, 8, 64) \\
    Projection 2   &  \texttt{Linear}                          & (20, 80, 80, 8, 1) \\
    \bottomrule
  \end{tabular}
  \label{table:postinjfno}
      \end{minipage}
\end{center}
\end{table}

\begin{table}[h]
\begin{center}
\begin{minipage}{\textwidth}
  \centering
  \caption{Model parameters for CO$_2$ mass during the post-injection period. After the flattening layer, the model splits into three channels where each channel contains an FC1 and FC2.}\label{model_4}
  \footnotesize
  \begin{tabular}{@{}lll@{}}
    \toprule
    Layer     & Operation     & Output Shape \\
    \midrule
    Input & -                         & (20, 80, 80, 8) \\
    Conv1 & \texttt{Conv3D/BN/ReLU}   & (20, 40, 40, 4) \\
    Conv2 & \texttt{Conv3D/BN/ReLU}   & (20, 20, 20, 2)  \\
    Flatten & - & (1, 16000) \\
    FC1 & \texttt{FC/BN/ReLU}   & (1, 100) \\
    FC2 & \texttt{FC/BN/ReLU}   & (1, 20) \\
    \bottomrule
  \end{tabular}
  \label{table:postinjmass}
      \end{minipage}
\end{center}
\end{table}

\end{document}